\documentclass[11pt,a4paper]{article}
\usepackage[hyperref]{acl2017}
\usepackage{amsmath}
\usepackage{cleveref}

\usepackage{times}
\usepackage{latexsym}
\usepackage{url}

\usepackage{graphicx}
\usepackage{lipsum}
\usepackage{multicol}
\usepackage{graphicx}
\usepackage{mathrsfs}
\usepackage{multirow}
\usepackage{caption}
\usepackage{amssymb}

\usepackage[utf8]{inputenc}
\usepackage[english]{babel}
 
\aclfinalcopy 
\def\aclpaperid{699} 

\title{Deep Keyphrase Generation}

\author{Rui Meng, Sanqiang Zhao, Shuguang Han, Daqing He\thanks{Corresponding author}  , Peter Brusilovsky, Yu Chi \\
  School of Computing and Information\\
  University of Pittsburgh\\
  Pittsburgh, PA, 15213\\
  {\{rui.meng, saz31, shh69, daqing, peterb, yuc73\}@pitt.edu} 
}
\date{}

\begin{document}
\maketitle
\begin{abstract}
Keyphrase provides highly-condensed information that can be effectively used for understanding, organizing and retrieving text content. Though previous studies have provided many workable solutions for automated keyphrase extraction, they commonly divided the to-be-summarized content into multiple text chunks, then ranked and selected the most meaningful ones. These approaches could neither identify keyphrases that do not appear in the text, nor capture the real semantic meaning behind the text. We propose a generative model for keyphrase prediction with an encoder-decoder framework, which can effectively overcome the above drawbacks.  We name it as \textit{deep keyphrase generation} since it attempts to capture the deep semantic meaning of the content with a deep learning method. Empirical analysis on six datasets demonstrates that our proposed model not only achieves a significant performance boost on extracting keyphrases that appear in the source text, but also can generate absent keyphrases based on the semantic meaning of the text. Code and dataset are available at \href{https://github.com/memray/OpenNMT-kpg-release}{https://github.com/memray/OpenNMT-kpg-release}.
\end{abstract}

\section{Introduction}
A keyphrase or keyword is a short piece of text that summarizes the main semantic meaning of a longer text. The typical use of a keyphrase or keyword is in scientific publications to provide the core information of a paper. We use the term ``keyphrase'' interchangeably with ``keyword'' in the rest of this paper, as both terms have an implication that they may contain multiple words. High-quality keyphrases can facilitate the understanding, organizing, and accessing of document content. As a result, many studies have focused on ways of automatically extracting keyphrases from textual content \cite{liu2009termclustering,medelyan2009maui,witten1999kea}. Due to public accessibility, many scientific publication datasets are often used as test beds for keyphrase extraction algorithms. Therefore, this study also focuses on extracting keyphrases from scientific publications.

Automatically extracting keyphrases from a document is called \textit{keypharase extraction}, and it has been widely used in many applications, such as information retrieval \cite{jones1999phrasier}, text summarization \cite{zhang2004world}, text categorization \cite{hulth2006study}, and opinion mining \cite{berend2011opinion}. Most of the existing keyphrase extraction algorithms have addressed this problem through two steps \cite{liu2009termclustering,Tomokiyo:2003:languagemodel}. The first step is to acquire a list of keyphrase candidates. Researchers have tried to use n-grams or noun phrases with certain part-of-speech patterns for identifying potential candidates \cite{hulth2003improved,le2016unsupervised,liu2010topicdecomposition,Wang2016PTR}. The second step is to rank candidates on their importance to the document, either through supervised or unsupervised machine learning methods with a set of manually-defined features \cite{frank1999domain,liu2009termclustering,liu2010topicdecomposition,kelleher2005automatic,matsuo2004keyword,mihalcea2004textrank,song2003kpspotter,witten1999kea}. 

There are two major drawbacks in the above keyphrase extraction approaches. First, these methods can only extract the keyphrases that appear in the source text; they fail at predicting meaningful keyphrases with a slightly different sequential order or those that use synonyms. However, authors of scientific publications commonly assign keyphrases based on their semantic meaning, instead of following the written content in the publication. In this paper, we denote phrases that do not match any contiguous subsequence of source text as \textbf{absent keyphrases}, and the ones that fully match a part of the text as \textbf{present keyphrases}. Table \ref{tab:ratio-present-absent} shows the proportion of present and absent keyphrases from the document abstract in four commonly-used datasets, from which we can observe large portions of absent keyphrases in all the datasets. The absent keyphrases cannot be extracted through previous approaches, which further prompts the development of a more powerful keyphrase prediction model.

Second, when ranking phrase candidates, previous approaches often adopted machine learning features such as TF-IDF and PageRank. However, these features only target to detect the importance of each word in the document based on the statistics of word occurrence and co-occurrence, and are unable to reveal the full semantics that underlie the document content.

\begin{table}[!htbp]
  \caption{Proportion of the present keyphrases and absent keyphrases in four public datasets}
  \label{tab:ratio-present-absent} 
  \fontsize{10}{10}\selectfont
  \renewcommand{\arraystretch}{1.4}

  \centering
  \begin{tabular}{cccc}
    \hline
    \hline
    \textbf{Dataset}
    & \textbf{\# Keyphrase}
    & \textbf{\% Present}
    & \textbf{\% Absent}
    \\    
    \hline
    \textbf{Inspec} & 19,275 & 55.69 & 44.31 \\  
    \hline
    \textbf{Krapivin} & 2,461 & 44.74 & 52.26 \\  
    \hline
    \textbf{NUS} & 2,834 & 67.75 & 32.25 \\  
    \hline
    \textbf{SemEval} & 12,296 & 42.01 & 57.99 \\  
    \hline
    \hline
  \end{tabular} 
\end{table}

To overcome the limitations of previous studies, we re-examine the process of \textit{keyphrase prediction} with a focus on how real human annotators would assign keyphrases. Given a document, human annotators will first read the text to get a basic understanding of the content, then they try to digest its essential content and summarize it into keyphrases. Their generation of keyphrases relies on an understanding of the content, which may not necessarily use the exact words that occur in the source text. For example, when human annotators see ``\textit{Latent Dirichlet Allocation}'' in the text, they might write down ``\textit{topic modeling}'' and/or ``\textit{text mining}'' as possible keyphrases. In addition to the semantic understanding, human annotators might also go back and pick up the most important parts, based on syntactic features. For example, the phrases following ``\textit{we propose/apply/use}'' could be important in the text. As a result, a better keyphrase prediction model should understand the semantic meaning of the content, as well as capture the contextual features.

To effectively capture both the semantic and syntactic features, we use recurrent neural networks (RNN) \cite{cho2014learning,gers2001lstm} to compress the semantic information in the given text into a dense vector (i.e., semantic understanding). Furthermore, we incorporate a copying mechanism \cite{gu2016copynet} to allow our model to find important parts based on positional information. Thus, our model can generate keyphrases based on an understanding of the text, regardless of the presence or absence of keyphrases in the text; at the same time, it does not lose important in-text information.

The contribution of this paper is three-fold. First, we propose to apply an RNN-based generative model to keyphrase prediction, as well as  incorporate a copying mechanism in RNN, which enables the model to successfully predict phrases that rarely occur. Second, this is the first work that concerns the problem of absent keyphrase prediction for scientific publications, and our model recalls up to 20\% of absent keyphrases. Third, we conducted a comprehensive comparison against six important baselines on a broad range of datasets, and the results show that our proposed model significantly outperforms existing supervised and unsupervised extraction methods.


In the remainder of this paper, we first review the related work in Section \ref{sec::related-work}. Then, we elaborate upon the proposed model in Section \ref{sec::methodology}. After that, we present the experiment setting in Section \ref{sec::experiment-setting} and results in Section \ref{sec::results-and-analysis}, followed by our discussion in Section \ref{sec::discussion}. Section \ref{sec::conclusions} concludes the paper.

\section{Related Work}

\label{sec::related-work}
\subsection{Automatic Keyphrase Extraction}
A keyphrase provides a succinct and accurate way of describing a subject or a subtopic in a document. A number of extraction algorithms have been proposed, and the process of extracting keyphrases can typically be broken down into two steps. 

The first step is to generate a list of phrase candidates with heuristic methods. As these candidates are prepared for further filtering, a considerable number of candidates are produced in this step to increase the possibility that most of the correct keyphrases are kept.  The primary ways of extracting candidates include retaining word sequences that match certain part-of-speech tag patterns (e.g., nouns, adjectives) \cite{liu2011wam,Wang2016PTR,le2016unsupervised}, and extracting important n-grams or noun phrases \cite{hulth2003improved,medelyan2008topic}. 

The second step is to score each candidate phrase for its likelihood of being a keyphrase in the given document. The top-ranked candidates are returned as keyphrases. Both supervised and unsupervised machine learning methods are widely employed here. For supervised methods, this task is solved as a binary classification problem, and various types of learning methods and features have been explored \cite{frank1999domain,witten1999kea,hulth2003improved,Medelyan:2009:Maui,lopez2010humb,Gollapalli:2014:CeKE}. As for unsupervised approaches, primary ideas include finding the central nodes in text graph \cite{mihalcea2004textrank,Grineva:2009:EKT:1526709.1526798}, detecting representative phrases from topical clusters \cite{liu2009termclustering,liu2010topicdecomposition}, and so on.

Aside from the commonly adopted two-step process, another two previous studies realized the keyphrase extraction in entirely different ways. \newcite{Tomokiyo:2003:languagemodel} applied two language models to measure the phraseness and informativeness of phrases. \newcite{liu2011wam} share the most similar ideas to our work. They used a word alignment model, which learns a translation from the documents to the keyphrases. This approach alleviates the problem of vocabulary gaps between source and target to a certain degree. However, this translation model is unable to handle semantic meaning. Additionally, this model was trained with the target of title/summary to enlarge the number of training samples, which may diverge from the real objective of generating keyphrases.

\newcite{zhang-EtAl:2016:EMNLP20164} proposed a joint-layer recurrent neural network model to extract keyphrases from tweets, which is another application of deep neural networks in the context of keyphrase extraction. However, their work focused on sequence labeling, and is therefore not able to predict absent keyphrases.

\subsection{Encoder-Decoder Model}

The RNN Encoder-Decoder model (which is also referred as sequence-to-sequence Learning) is an end-to-end approach. It was first introduced by \newcite{cho2014learning} and \newcite{sutskever2014sequence} to solve translation problems. As it provides a powerful tool for modeling variable-length sequences in an end-to-end fashion, it fits many natural language processing tasks and can rapidly achieve great successes \cite{DBLPemnlpRushCW15,vinyals2015grammar,serban2016building}.

Different strategies have been explored to improve the performance of the Encoder-Decoder model. The attention mechanism \cite{bahdanau2014neural} is a soft alignment approach that allows the model to automatically locate the relevant input components. In order to make use of the important information in the source text, some studies sought ways to copy certain parts of content from the source text and paste them into the target text \cite{allamanis2016convolutionalattention,gu2016copynet,zeng2016readagaincopy}. A discrepancy exists between the optimizing objective during training and the metrics during evaluation. A few studies attempted to eliminate this discrepancy by incorporating new training algorithms \cite{marc2016sequenceleveltraining} or by modifying the optimizing objectives~\cite{shen2016minimumrisktraining}.

\section{Methodology}
\label{sec::methodology}

This section will introduce our proposed deep keyphrase generation method in detail. First, the task of keyphrase generation is defined, followed by an overview of how we apply the RNN Encoder-Decoder model. Details of the framework as well as the copying mechanism will be introduced in Sections \ref{sec::detail-encoder-decoder} and \ref{sec::copymechanism}.



\subsection{Problem Definition}

Given a keyphrase dataset that consists of $\mathbf{N}$ data samples, the i-th data sample $ (\mathbf{x^{(i)}}, \mathbf{p^{(i)}}) $ contains one source text $\mathbf{x^{(i)}}$, and $M_i$ target keyphrases $\mathbf{p^{(i)}=(p^{(i,1)},p^{(i,2)},\ldots,p^{(i,M_i)})}$. Both the source text $\mathbf{x^{(i)}}$ and keyphrase $\mathbf{p^{(i,j)}}$ are sequences of words:
$$
\mathbf{x^{(i)}} = x^{(i)}_1, x^{(i)}_2 ,\ldots, x^{(i)}_{L_{\mathbf{x^i}}}
$$
$$
\mathbf{p^{(i,j)}} = y_{1}^{(i,j)},y_{2}^{(i,j)},\ldots,y_{L_{\mathbf{p^{(i,j)}}}}^{(i,j)}
$$
$L_{\mathbf{x^{(i)}}}$ and $ L_{\mathbf{p^{(i,j)}}} $denotes the length of word sequence of $\mathbf{x^{(i)}}$ and $\mathbf{p^{(i,j)}}$ respectively.

Each data sample contains one source text sequence and multiple target phrase sequences. To apply the RNN Encoder-Decoder model, the data need to be converted into text-keyphrase pairs that contain only one source sequence and one target sequence. We adopt a simple way, which splits the data sample  $ (\mathbf{x^{(i)}}, \mathbf{p^{(i)}}) $ into $M_i$ pairs: $ (\mathbf{x^{(i)}},\mathbf{p^{(i,1)}}),(\mathbf{x^{(i)}},\mathbf{p^{(i,2)}}),\ldots, (\mathbf{x^{(i)}},\mathbf{p^{(i,M_i)}})$. Then the Encoder-Decoder model is ready to be applied to learn the mapping from the source sequence to target sequence. For the purpose of simplicity, $ (\mathbf{x}, \mathbf{y}) $ is used to denote each data pair in the rest of this section, where $\mathbf{x}$ is the word sequence of a source text and $\mathbf{y}$ is the word sequence of its keyphrase.


\subsection{Encoder-Decoder Model}

The basic idea of our keyphrase generation model is to compress the content of source text into a hidden representation with an encoder and to generate corresponding keyphrases with the decoder, based on the representation
. Both the encoder and decoder are implemented with recurrent neural networks (RNN).

The encoder RNN converts the variable-length input sequence $\mathbf{x}=(x_1,x_2,...,x_{T})$ into a set of hidden representation $ \mathbf{h} = (h_1, h_2, \ldots, h_T) $, by iterating the following equations along time $t$:

\begin{equation}
\label{eq:encoder}
\begin{aligned}
  &\mathbf{h}_t = \mathit{f}(x_t,\mathbf{h}_{t-1})\\
\end{aligned}
\end{equation}
where $f$ is a non-linear function. We get the context vector $\mathbf{c}$ acting as the representation of the whole input $\mathbf{x}$ through a non-linear function q.

\begin{equation}
\label{eq:contextvector}
\begin{aligned}
  &\mathbf{c} = q(h_1, h_2, ..., h_T)
\end{aligned}
\end{equation}

The decoder is another RNN; it decompresses the context vector and generates a variable-length sequence $\mathbf{y}=(y_1,y_2,...,y_{T'})$ word by word, through a conditional language model:
\begin{equation}
\label{eq:decoder}
\begin{aligned}
&\mathbf{s}_t=f(y_{t-1},\mathbf{s}_{t-1},\mathbf{c})
\\
&p(y_t|y_{1,...,t-1},\mathbf{x})=g(y_{t-1},\mathbf{s}_t,\mathbf{c})
\end{aligned}
\end{equation}
where $\mathbf{s}_t$ is the hidden state of the decoder RNN at time $ t $. The non-linear function $g$ is a softmax classifier, which outputs the probabilities of all the words in the vocabulary. $y_t$ is the predicted word at time $t$, by taking the word with largest probability after $g(\cdot)$.

The encoder and decoder networks are trained jointly to maximize the conditional probability of the target sequence, given a source sequence. After training, we use the beam search to generate phrases and a max heap is maintained to get the predicted word sequences with the highest probabilities.



\subsection{Details of the Encoder and Decoder}
\label{sec::detail-encoder-decoder}
A bidirectional gated recurrent unit (GRU) is applied as our encoder to replace the simple recurrent neural network. Previous studies \cite{bahdanau2014neural,cho2014learning} indicate that it can generally provide better performance of language modeling than a simple RNN and a simpler structure than other Long Short-Term Memory networks~\cite{hochreiter1997long}. As a result, the above non-linear function $f$ is replaced by the $\mathbf{GRU}$ function (see in \cite{cho2014learning}).
 
Another forward GRU is used as the decoder. In addition, an attention mechanism is adopted to improve performance. The attention mechanism was firstly introduced by \newcite{bahdanau2014neural} to make the model dynamically focus on the important parts in input. The context vector $\mathbf{c}$ is computed as a weighted sum of hidden representation $\mathbf{h}=(h_1,\ldots,h_{T})$:
\begin{equation}
\label{eq:attention}
\begin{aligned}
& \mathbf{c}_i = \sum_{j=1}^{T} \alpha_{ij}h_j
\\
& \alpha_{ij} = \frac{\exp(a(s_{i-1}, h_j))}{\sum^{T}_{k=1}\exp(a(s_{i-1},h_k))}
\end{aligned}
\end{equation}
where $a(s_{i-1}, h_j)$ is a soft alignment function that measures the similarity between $s_{i-1}$ and $h_j$; namely, to which degree the inputs around position $j$ and the output at position $i$ match. 

\subsection{Copying Mechanism}
\label{sec::copymechanism}
To ensure the quality of learned representation and reduce the size of the vocabulary, typically the RNN model considers a certain number of frequent words (e.g. 30,000 words in \cite{cho2014learning}), but a large amount of long-tail words are simply ignored. Therefore, the RNN is not able to recall any keyphrase that contains out-of-vocabulary words. Actually, important phrases can also be identified by positional and syntactic information in their contexts, even though their exact meanings are not known. The copying mechanism \cite{gu2016copynet} is one feasible solution that enables RNN to predict out-of-vocabulary words by selecting appropriate words from the source text.






By incorporating the copying mechanism, the probability of predicting each new word $y_t$ consists of two parts. The first term is the probability of generating the term (see Equation \ref{eq:decoder}) and the second one is the probability of copying it from the source text:
\begin{equation}
\label{eq:mixed_probability}
\begin{aligned}
 &p(y_t|y_{1,...,t-1}, \mathbf{x}) \\
 &=  p_{g}(y_t|y_{1,...,t-1}, \mathbf{x}) +  p_{c}(y_t|y_{1,...,t-1}, \mathbf{x})
\end{aligned}
\end{equation}



Similar to attention mechanism, the copying mechanism weights the importance of each word in source text with a measure of positional attention. But unlike the generative RNN which predicts the next word from all the words in vocabulary, the copying part $p_{c}(y_t|y_{1,...,t-1}, \mathbf{x})$ only considers the words in source text. Consequently, on the one hand, the RNN with copying mechanism is able to predict the words that are out of vocabulary but in the source text; on the other hand, the model would potentially give preference to the appearing words, which caters to the fact that most keyphrases tend to appear in the source text.
 \begin{equation}
 \label{eq:copy_score}
 \begin{aligned}
p_{c}(y_t|y_{1,...,t-1}, \mathbf{x})&=\frac{1}{Z}\sum_{j:x_j=y_t}\exp(\psi_c(x_j)), y\in\chi
\\
\psi_c(x_j) &= \sigma(\mathbf{h}_j^T\mathbf{W}_c)\mathbf{s}_t
 \end{aligned}
 \end{equation}
where $\chi$ is the set of all of the unique words in the source text $\mathbf{x}$, $\sigma$ is a non-linear function and $\mathbf{W_c} \in \mathbb{R}$ is a learned parameter matrix. $\mathbf{Z}$ is the sum of all the scores and is used for normalization. Please see \cite{gu2016copynet} for more details.

\section{Experiment Settings}
\label{sec::experiment-setting}
This section begins by discussing how we designed our evaluation experiments, followed by the description of training and testing datasets. Then, we introduce our evaluation metrics and baselines.

\subsection{Training Dataset}
\label{sec::training-dataset}
There are several publicly-available datasets for evaluating keyphrase generation. The largest one came from \newcite{krapivin2009largedataset}, which contains 2,304 scientific publications. However, this amount of data is unable to train a robust recurrent neural network model. In fact, there are millions of scientific papers available online, each of which contains the keyphrases that were assigned by their authors. Therefore, we collected a large amount of high-quality scientific metadata in the computer science domain from various online digital libraries, including ACM Digital Library, ScienceDirect, Wiley, and Web of Science etc.~\cite{han2013supporting,rui2016knowledge}. In total, we obtained a dataset of 567,830 articles, after removing duplicates and overlaps with testing datasets, which is 200 times larger than the one of \newcite{krapivin2009largedataset}. Note that our model is only trained on 527,830 articles, since 40,000 publications are randomly held out, among which 20,000 articles were used for building a new test dataset \textbf{KP20k}. Another 20,000 articles served as the validation dataset to check the convergence of our model, as well as the training dataset for supervised baselines.



\subsection{Testing Datasets}
For evaluating the proposed model more comprehensively, four widely-adopted scientific publication datasets were used. In addition, since these datasets only contain a few hundred or a few thousand publications, we contribute a new testing dataset \textbf{KP20k} with a much larger number of scientific articles. We take the title and abstract as the source text. Each dataset is described in detail below.

\begin{itemize}
\item[--] {\bf Inspec} \cite{hulth2003improved}: This dataset provides 2,000 paper abstracts. We adopt the 500 testing papers and their corresponding \textit{uncontrolled keyphrases} for evaluation, and the remaining 1,500 papers are used for training the supervised baseline models.


\item[--] {\bf Krapivin} \cite{krapivin2009largedataset}: This dataset provides 2,304 papers with full-text and author-assigned keyphrases. However, the author did not mention how to split testing data, so we selected the first 400 papers in alphabetical order as the testing data, and the remaining papers are used to train the supervised baselines.
\item[--] {\bf NUS} \cite{nguyen2007nus}: We use both \textit{author-assigned} and \textit{reader-assigned} keyphrases and treat all 211 papers as the testing data. Since the NUS dataset did not specifically mention the ways of splitting training and testing data, the results of the supervised baseline models are obtained through a five-fold cross-validation.
\item[--] {\bf SemEval-2010} \cite{kim2010semeval}: 288 articles were collected from the ACM Digital Library. 100 articles were used for testing and the rest were used for training supervised baselines.
\item[--] {\bf KP20k}: We built a new testing dataset that contains the titles, abstracts, and keyphrases of 20,000 scientific articles in computer science. They were randomly selected from our obtained 567,830 articles. Due to the memory limits of implementation, we were not able to train the supervised baselines on the whole training set. Thus we take the 20,000 articles in the validation set to train the supervised baselines. It is worth noting that we also examined their performance by enlarging the training dataset to 50,000 articles, but no significant improvement was observed.
\end{itemize}

\subsection{Implementation Details}
In total, there are 2,780,316 $\langle$text, keyphrase$\rangle$ pairs for training, in which text refers to the concatenation of the title and abstract of a publication, and keyphrase indicates an author-assigned keyword. The text pre-processing steps including tokenization, lowercasing and replacing all digits with symbol $\langle$digit$\rangle$ are applied. Two encoder-decoder models are trained, one with only attention mechanism (RNN) and one with both attention and copying mechanism enabled (CopyRNN). For both models, we choose the top 50,000 frequently-occurred words as our vocabulary, the dimension of embedding is set to 150, the dimension of hidden layers is set to 300, and the word embeddings are randomly initialized with uniform distribution in [-0.1,0.1]. Models are optimized using Adam \cite{kingma2014adam} with initial learning rate = $10^{-4}$, gradient clipping = 0.1 and dropout rate = 0.5. The max depth of beam search is set to 6, and the beam size is set to 200. The training is stopped once convergence is determined on the validation dataset (namely early-stopping, the cross-entropy loss stops dropping for several iterations).

In the generation of keyphrases, we find that the model tends to assign higher probabilities for shorter keyphrases, whereas most keyphrases contain more than two words. To resolve this problem, we apply a simple heuristic by preserving only the first single-word phrase (with the highest generating probability) and removing the rest. 

\subsection{Baseline Models}
Four unsupervised algorithms (Tf-Idf, TextRank~\cite{mihalcea2004textrank}, SingleRank~\cite{wan2008singlerank}, and ExpandRank~\cite{wan2008singlerank}) and two supervised algorithms (KEA~\cite{witten1999kea} and Maui~\cite{medelyan2009maui}) are adopted as baselines. We set up the four unsupervised methods following the optimal settings in \cite{hasan2010makesenseofstateoftheart}, and the two supervised methods following the default setting as specified in their papers.

\subsection{Evaluation Metric}
Three evaluation metrics, the macro-averaged \textit{precision}, \textit{recall} and \textit{F-measure} (F$_1$) are employed for measuring the algorithm's performance. Following the standard definition, precision is defined as the number of correctly-predicted keyphrases over the number of all predicted keyphrases, and recall is computed by the number of correctly-predicted keyphrases over the total number of data records. Note that, when determining the match of two keyphrases, we use Porter Stemmer for pre-processing.

\section{Results and Analysis}
\label{sec::results-and-analysis}
We conduct an empirical study on three different tasks to evaluate our model.

\begin{table*}[!htbp]
  \centering
  \fontsize{10}{12}\selectfont
  \renewcommand{\arraystretch}{1.2}
  \begin{tabular}{c|cc|cc|cc|cc|cc}
    \hline
    \hline
    \multirow{2}{*}{\textbf{Method}}  
    & \multicolumn{2}{c|}{\textbf{Inspec}}
    & \multicolumn{2}{c|}{\textbf{Krapivin}}
    & \multicolumn{2}{c|}{\textbf{NUS}}
    & \multicolumn{2}{c|}{\textbf{SemEval}}
    & \multicolumn{2}{c}{\textbf{KP20k}}
    \\
    &  \textbf{F$_1$@5}& \textbf{F$_1$@10} & \textbf{F$_1$@5} & \textbf{F$_1$@10}
    & \textbf{F$_1$@5} & \textbf{F$_1$@10} & \textbf{F$_1$@5} & \textbf{F$_1$@10} & \textbf{F$_1$@5} & \textbf{F$_1$@10}
    \\  
    \hline
    \hline
    Tf-Idf 
    & 0.223 & \underline{0.304}
    & 0.113 & 0.143
    & 0.139 & 0.181
    & 0.120 & \underline{0.184}
    & 0.105 & 0.130
    \\ 
    \hline
    TextRank 
    & \underline{0.229} & 0.275
    & 0.172 & 0.147
    & 0.195 & 0.190
    & \underline{0.172} & 0.181
    & 0.180 & 0.150
    \\ \hline
	SingleRank 
    & 0.214 & 0.297 
    & 0.096 & 0.137
    & 0.145 & 0.169
    & 0.132 & 0.169
    & 0.099 & 0.124
    \\ \hline
	ExpandRank 
    & 0.211 & 0.295
    & 0.096 & 0.136
    & 0.137 & 0.162
    & 0.135 & 0.163
    & N/A & N/A
    \\ \hline
	Maui 
    & 0.040 & 0.033 
    & \underline{0.243} & \underline{0.208}
    & \underline{0.249} & \underline{0.261}
    & 0.045 & 0.039
    & \underline{0.265} & \underline{0.227}
    \\  \hline
	KEA 
    & 0.109 & 0.129  
    & 0.096 & 0.136
    & 0.068 & 0.081
    & 0.027 & 0.027
    & 0.180 & 0.163
    \\  \hline\hline
	RNN 
    & 0.000 & 0.000 
    & 0.002 & 0.001
    & 0.005 & 0.004
    & 0.004 & 0.003
    & 0.138 & 0.009
    \\  \hline
	CopyRNN 
    & \textbf{0.292} & \textbf{0.336}
    & \textbf{0.302} & \textbf{0.252}
    & \textbf{0.342} & \textbf{0.317}
    & \textbf{0.291} & \textbf{0.296}
    & \textbf{0.328} & \textbf{0.255}
    \\  \hline
    \hline
  \end{tabular}
  \caption{The performance of predicting present keyphrases of various models on five benchmark datasets}
  \label{tab:performance-comparison}
\end{table*}

\subsection{Predicting Present Keyphrases}
This is the same as the keyphrase extraction task in prior studies, in which we analyze how well our proposed model performs on a commonly-defined task. To make a fair comparison, we only consider the present keyphrases for evaluation in this task. Table 2 provides the performances of the six baseline models, as well as our proposed models (i.e., RNN and CopyRNN). For each method, the table lists its F-measure at top 5 and top 10 predictions on the five datasets. The best scores are highlighted in bold and the underlines indicate the second best performances.

The results show that the four unsupervised models (Tf-idf, TextTank, SingleRank and ExpandRank) have a robust performance across different datasets. The ExpandRank fails to return any result on the KP20k dataset, due to its high time complexity. The measures on NUS and SemEval here are higher than the ones reported in \cite{hasan2010makesenseofstateoftheart} and \cite{kim2010semeval}, probably because we utilized the paper abstract instead of the full text for training, which may filter out some noisy information. The performance of the two supervised models (i.e., Maui and KEA) were unstable on some datasets, but Maui achieved the best performances on three datasets among all the baseline models.

As for our proposed keyphrase prediction approaches, the RNN model with the attention mechanism did not perform as well as we expected. It might be because the RNN model is only concerned with finding the hidden semantics behind the text, which may tend to generate keyphrases or words that are too general and may not necessarily refer to the source text. In addition, we observe that 2.5\% (70,891/2,780,316) of keyphrases in our dataset contain out-of-vocabulary words, which the RNN model is not able to recall, since the RNN model can only generate results with the 50,000 words in vocabulary. This indicates that a pure generative model may not fit the extraction task, and we need to further link back to the language usage within the source text. 
The CopyRNN model, by considering more contextual information, significantly outperforms not only the RNN model but also all baselines, exceeding the best baselines by more than 20\% on average. This result demonstrates the importance of source text to the extraction task. Besides, nearly 2\% of all correct predictions contained out-of-vocabulary words. 

\begin{figure*}[!htbp]
  \centering
  \includegraphics[width=1\textwidth]{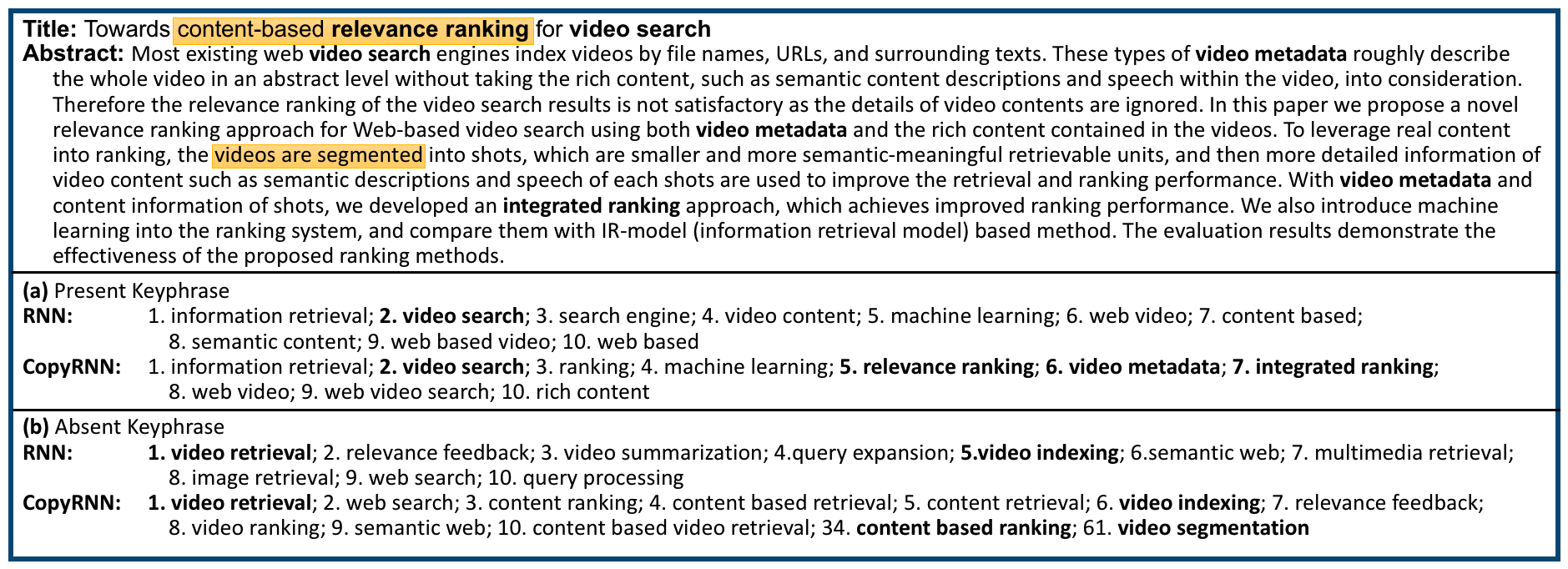}
  \caption{An example of predicted keyphrase by RNN and CopyRNN. Phrases shown in bold are correct predictions.}
  \label{fig:absent_keyphrase_sample}
\end{figure*}

The example in Figure \ref{fig:absent_keyphrase_sample}(a) shows the result of predicted present keyphrases by RNN and CopyRNN for an article about video search. We see that both models can generate phrases that relate to the topic of information retrieval and video. However most of RNN predictions are high-level terminologies, which are too general to be selected as keyphrases. CopyRNN, on the other hand, predicts more detailed phrases like ``video metadata'' and ``integrated ranking''. An interesting bad case, ``rich content'' coordinates with a keyphrase ``video metadata'', and the CopyRNN mistakenly puts it into prediction.

\subsection{Predicting Absent Keyphrases}
As stated, one important motivation for this work is that we are interested in the proposed model's capability for predicting absent keyphrases based on the ``understanding'' of content. It is worth noting that such prediction is a very challenging task, and, to the best of our knowledge, no existing methods can handle this task. Therefore, we only provide the RNN and CopyRNN performances in the discussion of the results of this task. Here, we evaluate the performance within the recall of the top 10 and top 50 results, to see how many absent keyphrases can be correctly predicted. We use the absent keyphrases in the testing datasets for evaluation. 

\begin{table}[!htbp]
  \centering
  \fontsize{10}{10}\selectfont
  \renewcommand{\arraystretch}{1.4}
  \begin{tabular}{c|cc|cc}
    \hline
    \hline
    \multirow{2}{*}{\textbf{Dataset}}  
    & \multicolumn{2}{c|}{\textbf{RNN}}
    & \multicolumn{2}{c}{\textbf{CopyRNN}}
    \\    
    & {\textbf{R@10}}
    & {\textbf{R@50}}
    & {\textbf{R@10}}
    & {\textbf{R@50}}
    \\
    \hline
    \textbf{Inspec} & 0.032 & 0.057 & \textbf{0.051} & \textbf{0.101}\\  
    \hline
    \textbf{Krapivin} & 0.095 & 0.155 & \textbf{0.116} & \textbf{0.195}\\  
    \hline
    \textbf{NUS} & 0.060 & 0.107 & \textbf{0.078} & \textbf{0.144}\\  
    \hline
    \textbf{SemEval} & 0.045 & 0.065 & \textbf{0.049} & \textbf{0.075}\\
    \hline
    \textbf{KP20k} & 0.076 & 0.129 & \textbf{0.115} & \textbf{0.189}\\  
    \hline
    \hline
  \end{tabular} 
  \caption{Absent keyphrases prediction performance of RNN and CopyRNN on five datasets}
\label{tab:predict-absent-keyphrases}
\end{table}

Table \ref{tab:predict-absent-keyphrases} presents the recall results of the top 10/50 predicted keyphrases for our RNN and CopyRNN models, in which we observe that the CopyRNN can, on average, recall around 8\% (15\%) of keyphrases at top 10 (50) predictions. This indicates that, to some extent, both models can capture the hidden semantics behind the textual content and make reasonable predictions. In addition, with the advantage of features from the source text, the CopyRNN model also outperforms the RNN model in this condition, though it does not show as much improvement as the present keyphrase extraction task. An example is shown in Figure \ref{fig:absent_keyphrase_sample}(b), in which we see that two absent keyphrases, ``video retrieval'' and ``video indexing'', are correctly recalled by both models. Note that the term ``indexing'' does not appear in the text, but the models may detect the information ``index videos'' in the first sentence and paraphrase it to the target phrase. And the CopyRNN successfully predicts another two keyphrases by capturing the detailed information from the text (highlighted text segments).





\subsection{Transferring the Model to the News Domain}
RNN and CopyRNN are supervised models, and they are trained on data in a specific domain and writing style. However, with sufficient training on a large-scale dataset, we expect the models to be able to learn universal language features that are also effective in other corpora. Thus in this task, we will test our model on another type of text, to see whether the model would work when being transferred to a different environment.

We use the popular news article dataset \textbf{DUC-2001}~\cite{wan2008singlerank} for analysis. The dataset consists of 308 news articles and 2,488 manually annotated keyphrases. The result of this analysis is shown in Table \ref{tab:performance_duc2001}, from which we could see that the CopyRNN can extract a portion of correct keyphrases from a unfamiliar text. We also report the baseline performance included in \cite{hasan2010makesenseofstateoftheart}. The performance of CopyRNN is better than TextRank~\cite{mihalcea2004textrank} and KeyCluster~\cite{liu2009termclustering}, but lags behind the other three baselines. It is worth noting that the hyperparameters of baseline models, such as number of recalled keyphrases for Tf-Idf and SingleRank, are carefully tuned and may drastically affect the results. However for CopyRNN we simply report its F${_1}$ score of top 10 predicted phrases (F${_1}$@10). 

As it is transferred to a corpus in a completely different type and domain, the model encounters more unknown words and has to rely more on the positional and syntactic features within the text. In this experiment, the CopyRNN recalls 766 keyphrases. 14.3\% of them contain out-of-vocabulary words, and many names of persons and places are correctly predicted.

\begin{table}[!htbp]
  \centering  
  \fontsize{10}{10}\selectfont
  \renewcommand{\arraystretch}{1.4}
  \begin{tabular}{l|c|l|c}
    \hline
    \hline
    {\textbf{Model}}
    & 
    {\textbf{F$_1$}}
    & 
    {\textbf{Model}}
    & 
    {\textbf{F$_1$}}
    
    \\    
    \hline
    \textbf{Tf-Idf} & 0.270 & \textbf{ExpandRank} & 0.269
    \\
    \textbf{TextRank} & 0.097 & \textbf{KeyCluster} & 0.140\\
    \textbf{SingleRank} & 0.256 & \textbf{CopyRNN} & 0.165 \\
    \hline
    \hline
  \end{tabular}
  \caption{Keyphrase prediction performance of CopyRNN on DUC-2001. The model is trained on scientific publication and evaluated on news.}
  \label{tab:performance_duc2001}
\end{table}

\section{Discussion}
\label{sec::discussion}
Our experimental results demonstrate that the CopyRNN model not only performs well on predicting present keyphrases, but also has the ability to generate topically relevant keyphrases that are absent in the text. In a broader sense, this model attempts to map a long text (i.e., paper abstract) with representative short text chunks (i.e., keyphrases), which can potentially be applied to improve information retrieval performance by generating high-quality index terms, as well as assisting user browsing by summarizing long documents into short, readable phrases.

Thus far, we have tested our model with scientific publications and news articles, and have demonstrated that our model has the ability to capture universal language patterns and extract key information from unfamiliar texts. We believe that our model has a greater potential to be generalized to other domains and types, like books, online reviews, etc., if it is trained on a larger data corpus. Also, we directly applied our model, which was trained on a publication dataset, into generating keyphrases for news articles without any adaptive training. 
We believe that with proper training on news data, the model would make further improvement.

Additionally, this work mainly studies the problem of discovering core content from textual materials. Here, the encoder-decoder framework is applied to model language; however, such a framework can also be extended to locate the core information on other data resources, such as  summarizing content from images and videos. 

\section{Conclusions and Future Work}
\label{sec::conclusions}
In this paper, we proposed an RNN-based generative model for predicting keyphrases in scientific text. To the best of our knowledge, this is the first application of the encoder-decoder model to a keyphrase prediction task. Our model summarizes phrases based the deep semantic meaning of the text, and is able to handle rarely-occurred phrases by incorporating a copying mechanism. 
Comprehensive empirical studies demonstrate the effectiveness of our proposed model for generating both present and absent keyphrases for different types of text. Our future work may include the following two directions.

\begin{itemize}
\item[--] In this work, we only evaluated the performance of the proposed model by conducting off-line experiments. In the future, we are interested in comparing the model to human annotators and using human judges to evaluate the quality of predicted phrases.

\item[--] Our current model does not fully consider correlation among target keyphrases. It would also be interesting to explore the multiple-output optimization aspects of our model.
\end{itemize}

\section*{Acknowledgments}
We would like to thank Jiatao Gu and Miltiadis Allamanis for sharing the source code and giving helpful advice. We also thank Wei Lu, Yong Huang, Qikai Cheng and other IRLAB members at Wuhan University for the assistance of dataset development. This work is partially supported by the National Science Foundation under Grant No.1525186.

\section*{Erratum}
We mistakenly reported the micro-averaged scores for all models instead of macro-averaged ones. We have updated all the scores to macro-averaged. As the difference between micro-averaged and macro-averaged score is marginal, this mistake doesn't affect any conclusions we drew in the submitted version. We sincerely apologize for this mistake and we thank Wang Chen from The Chinese University of Hong Kong for pointing it out.

\bibliographystyle{acl_natbib}
\bibliography{acl2017}

\end{document}